\documentclass[conference]{IEEEtran}
\IEEEoverridecommandlockouts
\usepackage{cite}
\usepackage{amsmath,amssymb,amsfonts}
\usepackage{algorithmic}
\usepackage{graphicx}
\usepackage{textcomp}
\usepackage[table]{xcolor}

\usepackage{multirow}
\usepackage{bbding}
\usepackage{pifont}
\usepackage{url}
\usepackage{amsmath}
\usepackage{booktabs}
\usepackage{amssymb}
\usepackage{bm}
\def\BibTeX{{\rm B\kern-.05em{\sc i\kern-.025em b}\kern-.08em
    T\kern-.1667em\lower.7ex\hbox{E}\kern-.125emX}}

\renewcommand{\footnoterule}{
\kern -3pt
\hrule width 0.2\textwidth height 0.5pt
\kern 2pt
}
    
\begin{document}

\title{DiffusionTalker: Efficient and Compact Speech-Driven 3D Talking Head via Personalizer-Guided Distillation}

\author{Peng Chen$^{1,2}$, Xiaobao Wei$^{1,2}$, Ming Lu$^{3}$, Hui Chen$^{1,2\dagger}$, Feng Tian$^{1,2}$ \\
$^1$Institute of Software, Chinese Academy of Sciences, Beijing, China \\
$^2$University of Chinese Academy of Sciences, Beijing, China \\
$^3$Intel Labs China, Beijing, China \\
chenpeng.cp0225@gmail.com
}

\maketitle

\renewcommand{\thefootnote}{\fnsymbol{footnote}} 
\footnotetext[2]{Corresponding Author.}

\begin{abstract}
Real-time speech-driven 3D facial animation has been attractive in academia and industry. Traditional methods mainly focus on learning a deterministic mapping from speech to animation. Recent approaches start to consider the nondeterministic fact of speech-driven 3D face animation and employ the diffusion model for the task.
Existing diffusion-based methods can improve the diversity of facial animation. However, personalized speaking styles conveying accurate lip language is still lacking, besides, efficiency and compactness still need to be improved.
In this work, we propose DiffusionTalker to address the above limitations via personalizer-guided distillation. 
In terms of personalization, we introduce a contrastive personalizer that learns identity and emotion embeddings to capture speaking styles from audio. We further propose a personalizer enhancer during distillation to enhance the influence of embeddings on facial animation.
For efficiency, we use iterative distillation to reduce the steps required for animation generation and achieve more than 8x speedup in inference.
To achieve compactness, we distill the large teacher model into a smaller student model, reducing our model's storage by 86.4\% while minimizing performance loss.
After distillation, users can derive their identity and emotion embeddings from audio to quickly create personalized animations that reflect specific speaking styles. Extensive experiments are conducted to demonstrate that our method outperforms state-of-the-art methods. The code will be released at: \url{https://github.com/ChenVoid/DiffusionTalker}.
\end{abstract}

\begin{IEEEkeywords}
Diffusion, Distillation, Efficiency, Compactness, Personalization
\end{IEEEkeywords}

\maketitle

\section{Introduction}
Speech-driven 3D facial animation is pivotal in applications such as virtual reality~\cite{wohlgenannt2020virtual, bai2024bring, chen2024mixedgaussianavatar, wei2024graphavatar, wei2024gazegaussian}, augmented reality~\cite{christoff2023application}, and computer games~\cite{ edwards2016jali}. This technology involves predicting facial parameters from audio sequences to drive a 3D facial model for animation. 
Real-time speech-driven 3D animation of special interests in communication anywhere, such as on mobile devices or VR glasses, which places high demands on the accuracy of lip movement, efficiency of inference speed and the compactness of model parameters.

With the advent of deep learning, data-driven techniques for speech-driven 3D animation have gained popularity~\cite{fan2022faceformer,fan2024unitalker,ma2024diffspeaker, haque2023facexhubert}, offering better performance.
VOCA~\cite{cudeiro2019capture} utilized CNNs for speech-driven 3D face generation with 12 FLAME-based facial models~\cite{li2017learning}, but ignored emotional speech impacts on expressions. EmoTalk~\cite{peng2023emotalk}, using Transformers~\cite{vaswani2017attention}, introduced an emotion disentangling encoder, but was limited by its reliance on user-defined one-hot identity encoding, restricting personalized speaking style capture.
SelfTalk~\cite{peng2023selftalk} uses a self-supervised approach, focusing on facial action units for better lip movement accuracy, but adds redundant encoders and decoders, increasing storage needs. 
Most methods employ deterministic networks, while speech-driven 3D facial animation is inherently nondeterministic~\cite{stan2023facediffuser}. FaceDiffuser~\cite{stan2023facediffuser} addresses this with a diffusion architecture, yielding impressive results, but its inference speed is slowed by a denoising process that requires thousands of steps.

\begin{figure*}
    \centering
    \includegraphics[width=0.9\linewidth]{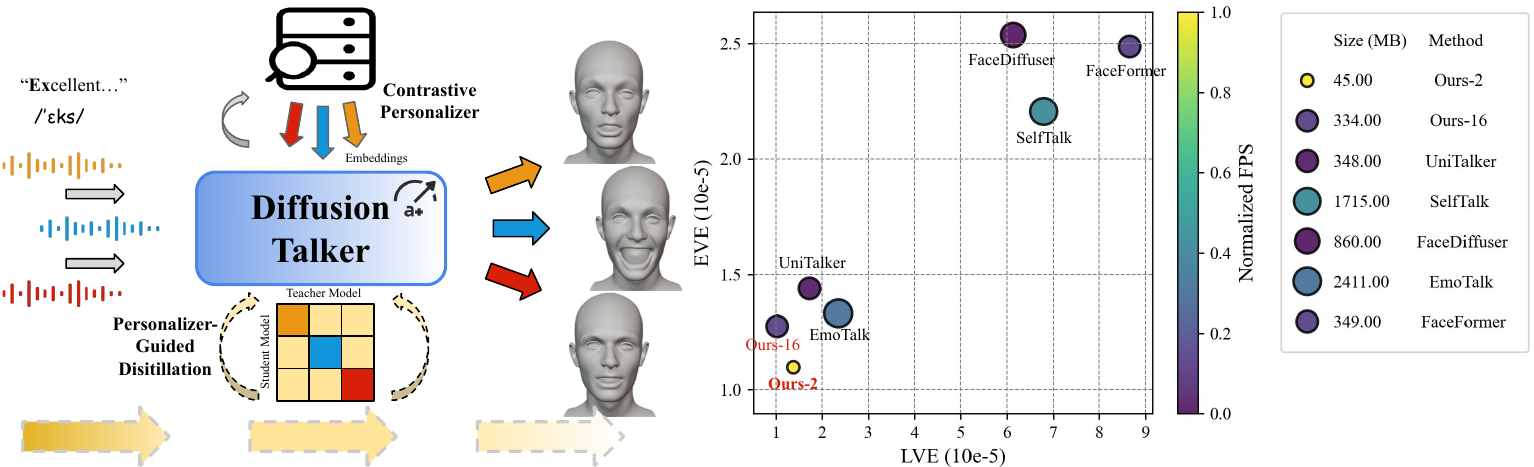}
    \vspace{-2mm}
    \caption{\textbf{The illustration of DiffusionTalker.} We reduce the steps of the diffusion model for faster inference and compress the model size for compactness by personalizer-guided distillation. Our distilled 2-step model surpasses the state-of-the-art methods in terms of emotional expression and lip accuracy, while also achieving the fastest inference speed and the fewest model parameters.}
    \vspace{-7mm}
    \label{fig:teaser}
\end{figure*}

Overall, current methods face three primary limitations.
First, the slow inference speed of existing approaches is a significant challenge, particularly for real-time applications, with long-step diffusion-based methods being especially impacted.
Second, many methods rely on large pre-trained audio encoders, such as HuBERT~\cite{hsu2021hubert} and Wav2Vec~\cite{baevski2020wav2vec}, which introduces redundancy in model parameters and leads to excessive storage requirements.
Third, most methods struggle to effectively integrate identity and emotion in 3D face animation. In practice, even when different people say the same words with the same emotions, their facial movements remain unique. And a person's expressions can vary greatly with their emotional state, even when repeating the same phrases.

To address existing limitations, we propose DiffusionTalker, an efficient and compact model that combines identity and emotion information to generate personalized 3D facial animations from speech, as illustrated in Fig.~\ref{fig:teaser}.
A contrastive personalizer module is proposed, in which identity and emotion are extracted from the audio input through contrastive learning. Using a cross-attention mechanism, these embeddings are fused into a personalized embedding, which serves as a key conditioning factor for the denoising process. This personalized embedding guides the motion decoder in generating customized facial animations.
To boost efficiency, a personalizer-guided distillation approach is used to iteratively distill a teacher model with N steps into a student model with n steps $(N > n)$, thereby accelerating the inference speed.
For model compression, we distill the large pre-trained audio encoder from the teacher model into a smaller encoder in the student model, significantly reducing both model parameters and storage requirements.

Our main contributions are summarized as follows.
\begin{itemize}
\item We introduce DiffusionTalker, an efficient and compact 3D face diffuser that generates personalized 3D facial animations based on the diffusion model.
\item A personalizer-guided distillation method is employed to iteratively reduce the denoising steps, significantly accelerating inference and achieving a speed-up of more than 8x. Furthermore, we distill a large model into a smaller one, resulting in 86.4\% reduction in model size.
\item We introduce a contrastive personalizer that learns identity and emotion embeddings using contrastive learning, which are then integrated into a personalized embedding through a cross-attention mechanism. 
\item A personalizer enhancer is applied during distillation to bring personalized embeddings of the same identity-emotion closer together and to push those of different identity-emotions farther apart, further enhancing the model's ability to generate personalized facial animations.
\end{itemize}

\begin{figure*}[h]
\vspace{-12pt}
    \centering
    \includegraphics[width=0.9\textwidth]{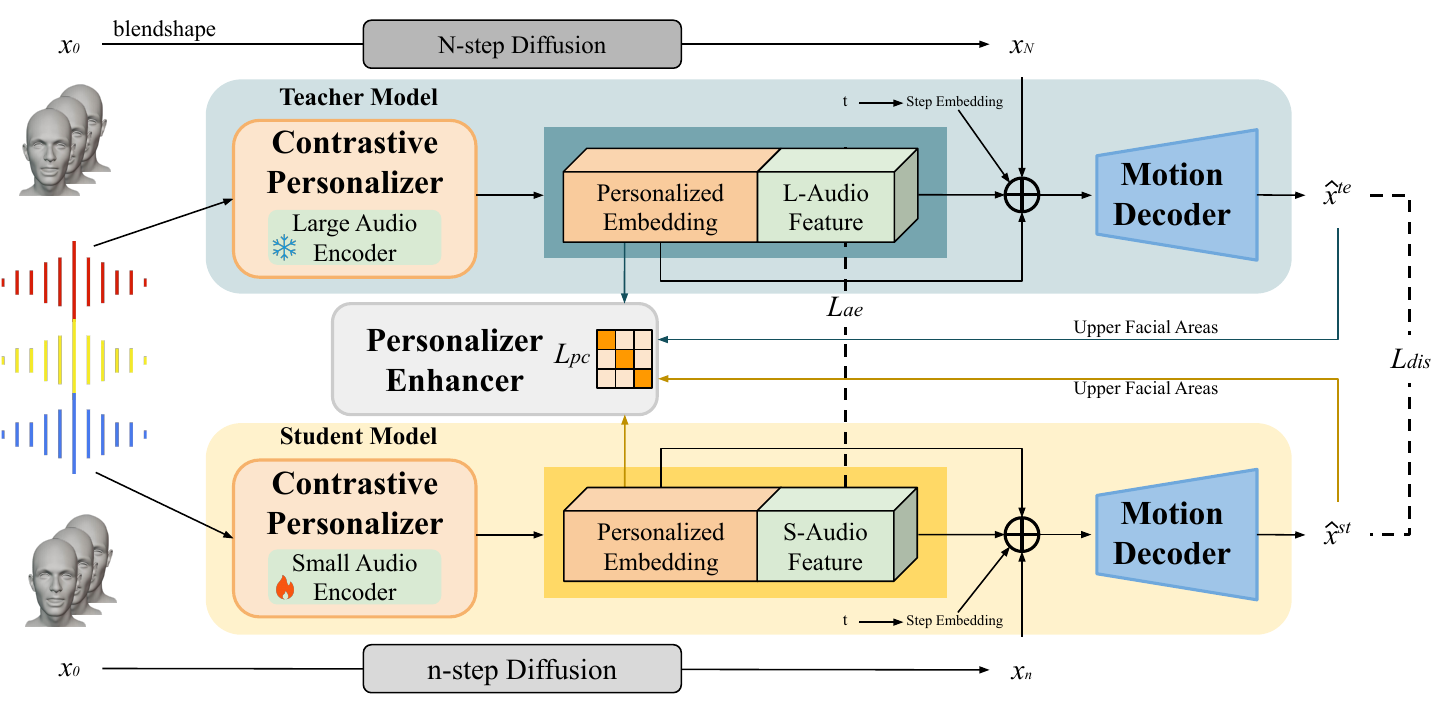}
    \vspace{-12pt}
    \caption{\textbf{Pipeline of DiffusionTalker.}
    DiffusionTalker employs a contrastive personalizer to extract audio features and personalized embeddings from the input speech. These representations serve as conditioning inputs to guide the motion decoder in denoising noisy facial animations effectively.  
    In personalizer-guided distillation process, the number of steps in the student model is iteratively reduced to half of the original, significantly accelerating inference. Simultaneously, the model parameters are compressed to create a more compact and efficient model. The personalizer enhancer integrates the personalized embedding with the facial areas of the predicted results, leveraging contrastive learning to strengthen the embedding’s representational capacity.}
    \label{fig:method}
    \vspace{-16pt}
\end{figure*}
\section{Related Work}

\textbf{Speech-driven 3D Facial Animation.} 
Methods for speech-driven 3D facial animation can be divided into two branches: phoneme-based methods~\cite{charalambous2019audio} and data-driven based methods~\cite{pham2018end, taylor2017deep, huang2018visual}. Phoneme-based methods, such as JALI~\cite{edwards2016jali, zhou2018visemenet}, utilize the mapping from phonemes to visemes and come with the advantage of simplified animation control~\cite{stan2023facediffuser}. However, they require intermediary representations of phonemes for co-articulation. 
Unlike phoneme-based methods, data-based methods can learn the mapping between audio and facial animation automatically. 
VOCA ~\cite{cudeiro2019capture} uses CNN to map voice to 3D mesh. While it considers the influence of different individuals' identities on animation generation, it does not account for emotions.
MeshTalk~\cite{richard2021meshtalk} attempts to disentangle audio and 3D animation in a categorical latent space but suffers from long inference times. FaceFormer~\cite{fan2022faceformer} employs a transformer as its model backbone and utilizes the attention mechanism to extract audio context features.
UniTalker~\cite{fan2024unitalker} employs various training strategies, such as Principal Component Analysis (PCA), model warm-up, and key identity embedding, to enhance training stability and ensure consistency across multi-head outputs.
EmoTalk~\cite{peng2023emotalk} concentrates on integrating emotion into FaceFormer to enhance facial animation using a cross-attention module~\cite{vaswani2017attention}, but its user-controlled method for identities cannot allow the model to learn the differences between personalized talking styles on its own.
FaceDiffuser~\cite{stan2023facediffuser} applies the diffusion model to this task, showcasing competitive generative capabilities and non-deterministic results. However, it necessitates 1000 steps of gradual denoising during the inference process, resulting in significant time consumption. 
The above methods suffer from slow inference speeds and redundant model parameters. And most of them only consider either identity or emotion individually, without effectively combining both aspects. 

\textbf{Personalization with Diffusion Model.} 
Personalization is a crucial but challenging technique in the research field of the diffusion model due to its diversity capability. 
DreamBooth~\cite{ruiz2023dreambooth} presents a novel pathway for the personalization of text-to-image diffusion models. 
Kumari et al.~\cite{kumari2023multi} demonstrate that optimizing a few parameters in diffusion models captures new concepts from images and proposes combining multiple concepts via constrained optimization. 
Liu et al.~\cite{liu2023cones} show that a small subset of neurons represents specific subjects, enabling the generation of related concepts within a single image. 
Han et al.~\cite{han2023svdiff} suggest fine-tuning weight matrix singular values for efficient personalization.
Van et al.~\cite{van2023anti} introduce subtle noise perturbations to safeguard against the misuse of personalization.
In our work, we integrate contrastive learning~\cite{radford2021learning} for the matching of identity and emotion with audio modalities into the diffusion model training to achieve personalization of facial animation based on input audio.
Kumari et al.~\cite{kumari2023multi} discover that optimizing a few parameters in diffusion models can represent new concepts from given images. They further suggest combining multiple concepts through closed-form constrained optimization.
Liu et al.~\cite{liu2023cones} find that a small subset of neurons can correspond to a specific subject, identified using network gradient statistics. By concatenating these concept neurons, they can generate all related concepts within a single image.
Han et al.~\cite{han2023svdiff} propose to fine-tune the singular values of the weight matrices, leading to a compact and efficient parameter space for personalization.
Van et al.~\cite{van2023anti} add subtle noise perturbations to the images for personalization before publishing to protect against malicious use of personalization.
A few recent works study the protection against malicious use of personalization techniques. For example, \cite{van2023anti} adds subtle noise perturbations to the images for personalization before publishing. 
However, for speech-driven 3D facial animation synthesis, there has been no related work using diffusion models for personalization.

\section{Methodology}

\vspace{-2mm}
\subsection{Preliminaries}
The vanilla Denoising Diffusion Probabilistic Model (DDPM)~\cite{ho2020denoising} consists of two processes: the diffusion process and the denoising process. The diffusion process gradually adds Gaussian noise to clean data \( \bm{x}_0 \) using a Markov chain, resulting in a noisy distribution \( q(\bm{x}_T | \bm{x}_0) \):
\begin{equation}
    q(\bm{x}_T|\bm{x}_0) = \prod_{t=1}^{T} q(\bm{x}_t|\bm{x}_{t-1})
\end{equation}
where \( T \) is the diffusion step. 
Specifically, the clean data \( \bm{x}_0 \) is noised to \( \bm{x}_t \), as shown in the following equation:
\begin{equation}
    \bm{x}_t = \sqrt{\bar{\alpha_t}} \bm{x}_0 + \sqrt{1 - \bar{\alpha_t}} \bm{\epsilon}
\end{equation}
where $\overline{\alpha_t}$ is hyperparameters, and $\bm{\epsilon}$ is Gaussian noise.

Denoising is an inverse stepwise process $q'(\bm{x}_{t-1}|\bm{x}_t)$, transitioning from \( \bm{x}_t \) to \( \bm{x}_{t-1} \), as shown in the following equation:
\begin{equation}
\bm{x}_{t-1} = \frac{1}{\sqrt{\alpha_t}} \left( \bm{x}_t - \frac{1 - \alpha_t}{\sqrt{1 - \bar{\alpha_t}}} \bm{\epsilon}_\theta(\bm{x}_t, t) \right) + \sigma_t \mathbf{z}
\end{equation}
where $\bm{\epsilon}_\theta$ is the network to predict noise, $\theta$ is the parameters, and $\sigma_t \mathbf{z}$ is a noise term used to introduce diversity. Similarly, the denoising process also involves \( T \) progressive steps.

\subsection{Overview}

DiffusionTalker uses a DDPM model, conditioning on audio, identity, and emotion to guide the denoising process for generating speech-driven 3D facial animations. As shown in Fig.~\ref{fig:method}, the model consists of the contrastive personalizer and the motion decoder. The personalizer extracts audio features, identity, and emotion embeddings from the input speech, combining them into a personalized embedding. The motion decoder uses these embeddings, along with the noisy facial animation \(\bm{x}_t\) and time step \(t\), to iteratively remove noise and predict the final facial animation \(\hat{\bm{x}}\).
The model is trained by minimizing the reconstruction loss $\mathcal{L}_{rec}$ between the prediction and the ground truth, as shown in the following formula:

\vspace{-2mm}
\begin{equation}
     \mathcal{L}_{rec} = \left\| \bm{x}_0 - \text{DT}_\theta(\bm{a}, \bm{i}, \bm{e}, \bm{x}_t, t) \right\|^2
\end{equation}
where DT is our model, $\theta$ represents the model parameters, $\bm{a}$ is the audio sequence, $\bm{i}$ and $\bm{e}$ are the embeddings of identity and emotion, respectively, and $\bm{x}_0$ is the ground truth.

For personalizer-guided distillation, we use an $N$-step model as the teacher model and an $n$-step (where $N = 2n$) model as the student. This reduces inference time and compresses the audio encoder to optimize storage and performance, while maintaining lip accuracy. 
The personalizer enhancer further strengthens the representational power of the embeddings through contrastive learning.

\subsection{Contrastive Personalizer}

\begin{figure}[t]
  \centering
   \includegraphics[width=0.85\linewidth]{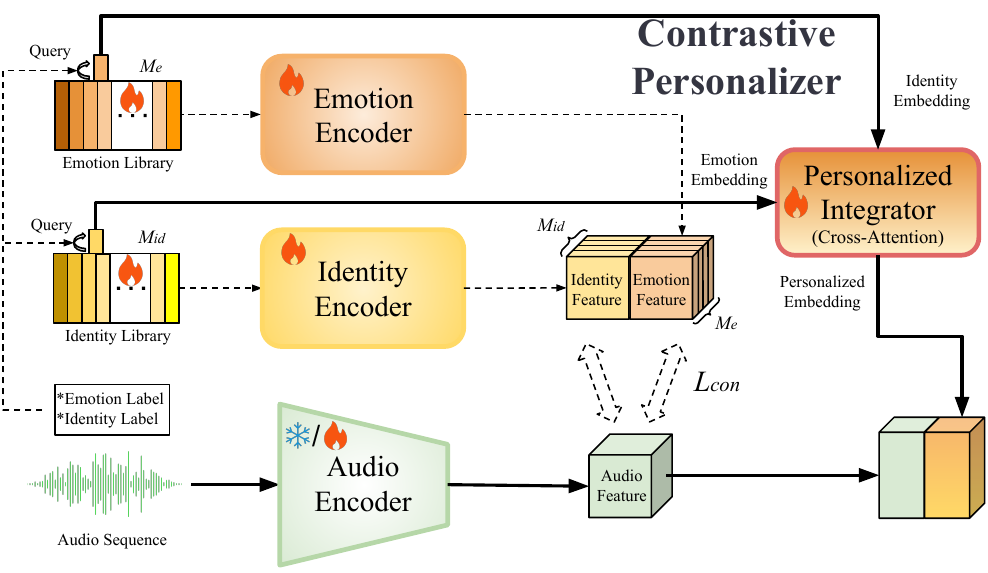}
   \vspace{-4mm}
   \caption{\textbf{Contrastive Personalizer.} Contrastive learning is performed separately on audio features and identity/emotion features, and specific identity/emotion embeddings are fused into personalized embeddings.}
   \label{fig:architecture}
   \vspace{-4mm}
\end{figure}

To equip the model with personalization ability, we propose a contrastive personalizer to extract and integrate identity and emotion from the input speech using contrastive learning similar to ~\cite{radford2021learning}.
As shown in Fig.~\ref{fig:architecture}, we establish an identity embedding library and an emotion library. Each speaker is associated with an identity one, $\bm{i}$, and each type of emotion corresponds to an emotion embedding, $\bm{e}$. 

The aim of this module is to formulate the relationship between identity/emotion embeddings and audio sequences. 
The audio sequence is first fed into the audio encoder for its feature, $\bm{F}_a$. Meanwhile, the libraries of identity and emotion are processed separately by the encoders of both to generate the corresponding features $\bm{F}_{id}$ and $\bm{F}_e$. These features are then used to calculate the contrastive loss with $F_a$ separately, which is adopted to train the parameters in this module.

Specifically, we first apply L2 normalization to both the audio feature and identity feature. Next, we perform matrix multiplication on these normalized features and the results are used to compute the cross-entropy loss with one-hot labels. The labels are constructed based on the positive or negative sample relationship between the audio feature and identity feature.         
For identity, $\bm{F}_a$ has only one positive sample, which is $\bm{F}_{id}$ that corresponds to the current identity label. 
All other embeddings are regarded as negative samples. The contrastive loss between $\bm{F}_{id}$ and $\bm{F}_a$ is formulated as:
\begin{equation} \label{eq:contrast}
     \mathcal{L}^{\bm{F}_{id}-\bm{F}_a}_{con} = -\log\frac{exp(\bm{F}_a^\top  \bm{F}_{id}^+/\tau)}{\sum_{k=1}^{M_{id}} exp(\bm{F}_a^\top \bm{F}_{id}^k/\tau)}
\end{equation}
where $M_{id}$ is the number of identity features, $\bm{F}^+$ is the positive sample of $\bm{F}_a$,  and $\tau$ is a temperature hyperparameter. 

For emotion, the formula of contrastive loss $L^{\bm{F}_e-\bm{F}_a}_{con}$ between $\bm{F}_e$ and $\bm{F}_a$ is similar to $L^{\bm{F}_{id}-\bm{F}_a}_{con}$, which is shown as:
\begin{equation} \label{eq:contrast}
     \mathcal{L}^{\bm{F}_{e}-\bm{F}_a}_{con} = -\log\frac{exp(\bm{F}_a^\top  \bm{F}_{e}^+/\tau)}{\sum_{k=1}^{M_e} exp(\bm{F}_a^\top \bm{F}_{e}^k/\tau)}
\end{equation}
where $M_e$ is the number of emotion features.

Simultaneously, the contrastive personalizer searches the libraries to retrieve $\bm{i}$ and $\bm{e}$ that match the current audio according to the input emotion and identity labels. $\bm{i}$ and $\bm{e}$ are then processed by the personalized integrator, which employs a cross-attention mechanism to merge them into a singular personalized embedding, $\bm{p}$. This composite embedding is then utilized as a key input for the decoder. The personalized integrator is illustrated as follows:
\vspace{-2mm}
\begin{equation}
    \bm{p} = \text{softmax}\left(\frac{\bm{i} \cdot f_k(\bm{e})^T}{\sqrt{d_k}}\right) \cdot f_v(\bm{e})
\end{equation}
where $\bm{i}$ is identity embedding, $\bm{e}$ is emotion embedding, $f_k$ and $f_v$ are MLPs, 
and $d_k$ is key dimension.

During inference, the personalizer can autonomously infer the emotion and identity embeddings that are most similar to the input audio sequence, without the need for labels.

The motion decoder processes a concatenated input consisting of the personalized embedding \(\bm{p}\), audio feature \(\bm{F}_a\), noisy animation \(\bm{x}_t\), and time step \(t\). It performs denoising from step \(t\) to step \(t-1\) to generate the predicted facial animation, represented as \(q'_m(\bm{x}_{t-1}|\bm{x}_t,\bm{F}_a,\bm{p},t)\).

To capture the temporal dynamics of speech, we use a GRU backbone, which is compact and efficient. The input is first normalized via layer normalization, then processed through three GRU layers to capture contextual information, and finally passed through an MLP to generate the facial animation.

\subsection{Personalizer-Guided Distillation}


\subsubsection{Acceleration}
To accelerate inference and ensure the high quality of the generated animations, we utilize this distillation manner for fewer denoising steps. 
Inspired by ~\cite{salimans2022progressive}, we introduce a student model $\hat{s}_{\eta}$ with $n$ steps to match the pre-trained teacher model $\hat{t}_{\theta}$ with $N$ steps (where $N=2n$). We use $\bm{c}_{t}=(\bm{p},\bm{F}_a, t)$ as the input of condition at time step $t$. Specifically, we perform one DDPM step for the student $n$ to match two DDPM steps for the teacher $2n$ and $2n-1$.

Assume that the signal-to-noise ratio at time step $\tau \sim \mathcal{U}(0, n)$ is $\alpha^2_\tau/\sigma^2_\tau$, then the corresponding data $\bm{x}_0$ added with noise $\bm{\epsilon}$ is $\bm{x}_\tau=\alpha_\tau \bm{x}_0 + \sigma_\tau \bm{\epsilon}, \bm{\epsilon} \sim N(0, I)$.

For the teacher model, we compute the denoised output of the $2n$ and $2n-1$ steps.
Thus, we can obtain noisy data $\bm{x}_{\tau'}$ at time step $\tau'=2\tau$ as: 
\vspace{-1mm}
\begin{equation}
    \hat{\bm{x}}_{\tau'} = \alpha_{\tau'}\hat{t}_{\theta}(\bm{x}_\tau, \bm{c}_{\tau'}) + \frac{\sigma_{\tau'}}{\sigma_{\tau}}(\bm{x}_\tau-\alpha_\tau\hat{t}_{\theta}(\bm{x}_\tau, \bm{c}_{\tau'}))
\end{equation}
At step $\tau''=2\tau-1$, noisy data $\bm{x}_{\tau''}$ can be calculated as:
\vspace{-1mm}
\begin{equation}
    \hat{\bm{x}}_{\tau''} = \alpha_{\tau''}\hat{t}_{\theta}(\hat{\bm{x}}_{\tau'}, \bm{c}_{\tau''}) + \frac{\sigma_{\tau''}}{\sigma_{\tau'}}(\hat{\bm{x}}_{\tau'}-\alpha_{\tau'} \hat{t}_{\theta}(\hat{\bm{x}}_{\tau'}, \bm{c}_{\tau''}))
\end{equation}
The final target $\hat{\bm{x}}^{te}$ from teacher model can be formulated as:
\vspace{-1mm}
\begin{equation}
    \hat{\bm{x}}^{te}=\frac{\hat{\bm{x}}_{\tau''}-(\sigma_{\tau''}/\sigma_{\tau})\bm{x}_\tau}{\alpha_{\tau''}-(\sigma_{\tau''}/\sigma_{\tau})\alpha_\tau}
\end{equation}

For the student model, we initialize it with the same parameters as the teacher. We then add noise according to the student DDPM for $\tau$ steps. This noisy data is then fed into the student, resulting in the output denoted as $\hat{\bm{x}}^{st}=\hat{s}_{\eta}(\bm{x}_{\tau},\bm{c}_\tau)$.

We apply $\hat{\bm{x}}^{te}$ to supervise the student model. The distill loss in this process is as follows:
\vspace{-2mm}
\begin{equation}
    \mathcal{L}_{dis} = max(\frac{\alpha^{2}_{\tau}}{\sigma^{2}_{\tau}}, 1)\|\hat{\bm{x}}^{te} - \hat{\bm{x}}^{st}\|^2 
\end{equation} 
Finally, DiffusionTalker can be distilled to half the original number of steps while preserving high generation quality.

\subsubsection{Compression}
During distillation, we assign a large pre-trained audio encoder to the teacher model to fully extract audio features, resulting in the large audio feature $\bm{F}^L_a$. For the student model, we assign a small audio encoder to obtain the small audio feature $\bm{F}^S_a$. We use an audio encoder (AE) loss based on cosine similarity to optimize the small audio encoder, aligning $\bm{F}^S_a$ with $\bm{F}^L_a$. The formula is as follows:

\begin{equation}
\mathcal{L}_{\text{ae}} = 1 - \text{mean} \left( \frac{\bm{F}^L_a}{\|\bm{F}^L_a\|_2} \cdot \frac{\bm{F}^S_a}{\|\bm{F}^S_a\|_2} \right)
\end{equation}

The large audio encoder employs the pre-trained HuBERT model~\cite{hsu2021hubert}, while the small one utilizes the feature extractor (CNN) from HuBERT to process audio input. Subsequently, it extracts temporal features through a bidirectional LSTM and maps them to the target dimension using a fully connected layer. Finally, the distilled student model significantly reduces model size while maintaining lip accuracy.

\begin{figure}[t]
  \centering
   \includegraphics[width=0.9\linewidth]{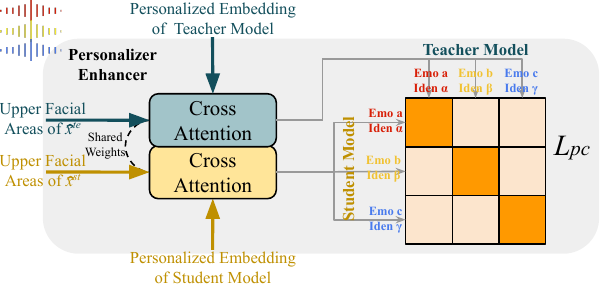}
   \vspace{-3mm}
   \caption{\textbf{Personalizer Enhancer.} It is used to enhance the personalization.}
   \label{fig:enhancer}
   \vspace{-6mm}
\end{figure}

\begin{figure*}[h]
  \centering
   \includegraphics[width=1.0\textwidth]{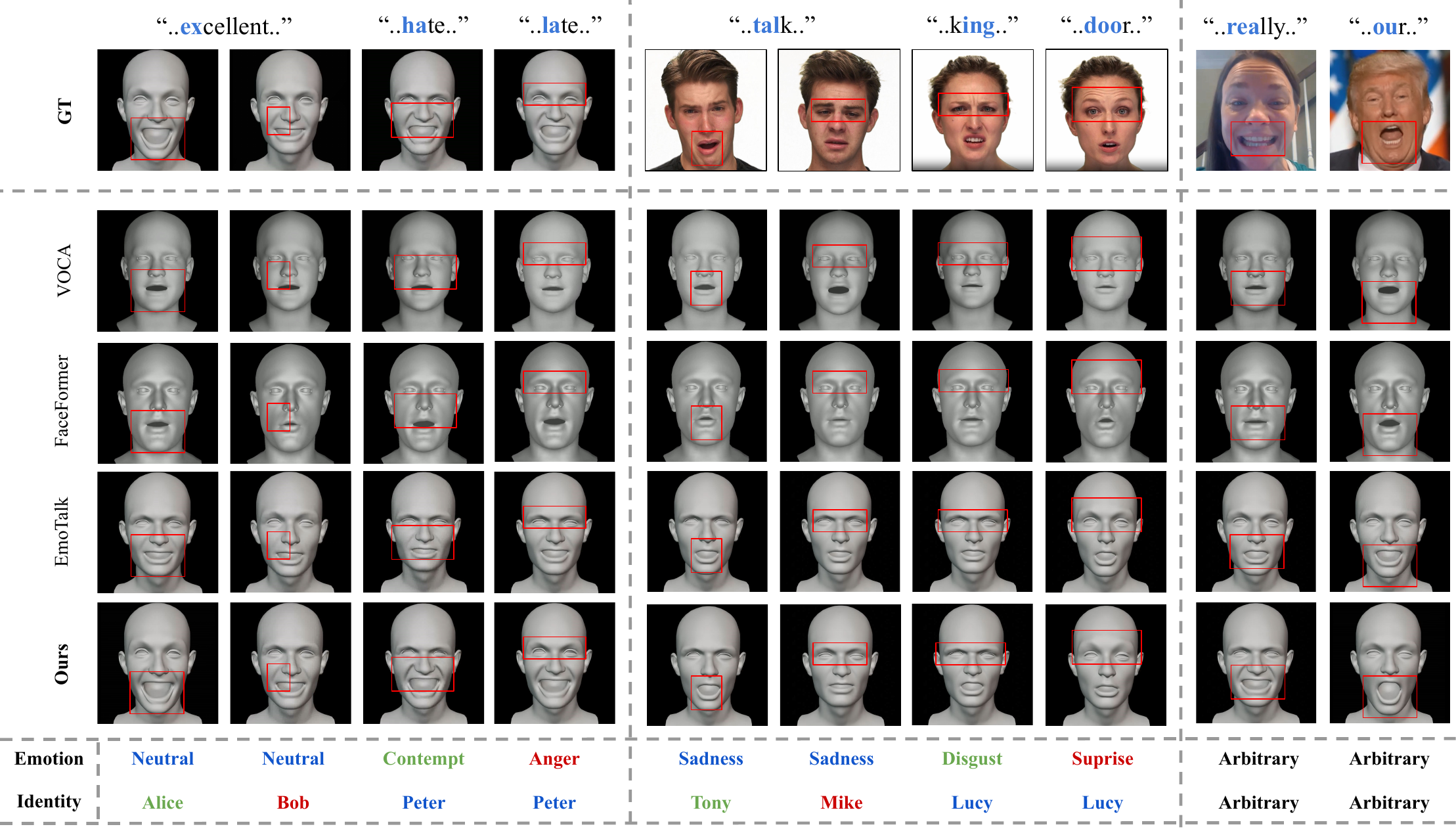}
   \vspace{-4mm}
   \caption{Qualitative comparisons with other methods on BEAT-Test(left), 3D-ETF(middle), and in-the-wild videos(right). We input speech with different identities and emotions into various models and present the same frames to compare them with the ground truth (GT). As indicated by the red box, it can be witnessed that on the first two datasets, our model accurately discerns facial action changes among various identities and precisely generates facial expressions corresponding to specific emotions. Even on in-the-wild videos, our model can produce accurate results.}
   \label{fig:visualization}
   \vspace{-6mm}
\end{figure*}

\subsubsection{Personalizer Enhancer}

To further enhance the influence of identity and emotion on the mapping from speech to facial animation, we propose a personalizer enhancer during the distillation process.

As illustrated in Fig.~\ref{fig:enhancer}, we set the batch size as \( M = 3 \). Specifically, the red speech icon represents an audio sequence corresponding to identity \( a \) with emotion \( \alpha \), the yellow icon denotes identity \( b \) with emotion \( \beta \), and the blue icon represents identity \( c \) with emotion \( \gamma \), where $\alpha \neq \beta \neq \gamma, a \neq b \neq c$.
In the context of the teacher model, the audio sequences are passed through the personalizer that generates three distinct personalized embeddings. Each personalized embedding \(\bm{p}\) is then employed in cross-attention with the upper facial areas of the predicted results \(\hat{\bm{x}}^{te}_u\) , resulting in three personalized feature maps \( \bm{F}^{te}_{pu} \), for the teacher model, where $f(\cdot)$ is MLP.
\vspace{-2mm}
\begin{equation}
    \bm{F}^{te}_{pu} = \text{softmax}\left(\frac{\bm{p} \cdot f_k(\hat{\bm{x}}^{te}_u)^T}{\sqrt{d_k}}\right) \cdot f_v(\hat{\bm{x}}^{te}_u)
\end{equation}
\vspace{-2mm}

A similar procedure is applied to the student model, yielding three corresponding feature maps \( \bm{F}^{st}_{pu} \).
The personalized feature maps \( \bm{F}^{te}_{pu} \) and \( \bm{F}^{st}_{pu} \) are then compared using a contrastive learning method, which computes the personalized contrastive (PC) loss \( \mathcal{L}_{pc} \). The formula is shown as:
\vspace{-2mm}
\begin{equation}
     \mathcal{L}_{pc} = -\frac{1}{M}\sum_{i=1}^M\log\frac{exp(\bm{F}^{te\;i\;\top}_{pu}  \bm{F}^{st\;+}_{pu}/\tau)}{\sum_{j=1}^M exp(\bm{F}^{te \;i\;\top}_{pu} \bm{F}^{st\;j}_{pu}/\tau)}
\end{equation}
\vspace{-2mm}

Notably, the teacher and student models share identical weight parameters for the cross-attention mechanism. More details are provided in the supplementary materials.

This approach effectively amplifies the contribution of personalized embeddings to the facial animation output, thereby improving the model’s capacity to capture and reflect identity and emotion in the generated animation.

\vspace{-2mm}
\subsection{Training}

The training process begins by using \( \mathcal{L}_{tea} \) to train an initial teacher model until convergence. The formula is shown as:
\vspace{-2pt}
\begin{equation}
    \mathcal{L}_{tea} = \lambda_{1}\mathcal{L}_{rec}+\lambda_{2}\mathcal{L}^{\bm{F}_{i}-\bm{F}_a}_{con}+\lambda_{2}\mathcal{L}^{\bm{F}_{e}-\bm{F}_a}_{con}
\end{equation}

This teacher model is then used for distillation to train a half-step student model using \( \mathcal{L}_{stu} \). The student model is subsequently promoted to the role of the teacher, facilitating iterative distillation to train a new half-step student model with \( \mathcal{L}_{stu} \), which is shown as:
\begin{equation}
    \mathcal{L}_{stu} = \mathcal{L}_{tea} +\lambda_{3}\mathcal{L}_{dis} +\lambda_{4}\mathcal{L}_{ae}+\lambda_{5}\mathcal{L}_{pc}
\end{equation}
where $\lambda_{1}$ = 1.0, $\lambda_{2}$ = 0.007, $\lambda_{3}$ = 0.1, $\lambda_{4}$ = 1, $\lambda_{5}$ = 0.05 are fixed in all our experiments. 

The trainable components of DiffusionTalker include the identity and emotion library, identity and emotion encoder, personalized integrator, small audio encoder, step embedding, motion decoder, and personalizer enhancer, while the parameters of the large audio encoder remain fixed. During distillation, the teacher's parameters are fixed, while the student's parameters are trainable.

\section{Experiments}

\vspace{-2mm}
\subsection{Datasets and Metrics}
To evaluate the robustness of DiffusionTalker, we select three datasets: BEAT~\cite{Liu2022beat}, 3D-ETF(RAVDESS)~\cite{peng2023emotalk}, and VOCASET~\cite{cudeiro2019capture}. Our method and the baselines are trained on the first two datasets. DiffusionTalker conducts zero-shot testing on the third dataset.
We follow the same data preprocessing as the baselines and use the EmoTalk's bs2FLAME~\cite{peng2023emotalk} method to align the data. 
Please refer to the supplementary materials for more details on these datasets.

To evaluate the performance of DiffusionTalker, we use five metrics: LVE, EVE, FDD, FPS, and model size.
\textbf{LVE}: Lip vertex error, measuring the maximum \(l_2\) error of the predicted vs. ground-truth lip areas.  
\textbf{EVE}: Emotional vertex error, assessing the maximum error \(l_2\) for the coefficients of the eyes, eyebrows, forehead, and surrounding areas. \textbf{FDD}: Facial dynamics deviation~\cite{xing2023codetalker}, quantifying upper-face dynamic differences; Lower values indicate more natural animations.  
\textbf{FPS}: Frames Per Second, indicating inference speed; higher values mean faster frame generation. \textbf{Model size} represents the number of parameters in the model.  

\subsection{Experimental Results}
\paragraph{Animation synthesis comparison.}
We test our DiffusionTalker model against other advanced methods on the 3D-ETF dataset with EVE, LVE, FDD, FPS and model size metrics, as shown in Tab.~\ref{tab:eval_on_3detf}. 
Our 16-step method (ours-16) serves as the initial teacher model and achieves the best performance across the LVE and FDD metrics, while our distilled 2-step model (ours-2) incurs minimal performance loss, achieving the second-best LVE and FDD results. Due to the effect of the personalizer enhancer, the EVE metric for emotion evaluation has decreased, indicating that the model’s ability to express personalization has become stronger.
Moreover, ours-2 achieves more than 8x speed-up in inference and an 86.4\% reduction in model size.
Compared to baselines, our 2-step model achieves the fastest speed and the smallest model size while delivering the best accuracy in lip region prediction (LVE). In terms of the naturalness of facial dynamic changes (FDD), our method also outperforms existing approaches.
Our method shows a significant advantage on the EVE metric, indicating that the personalized embeddings learned through the contrastive personalizer and personalizer-guided distillation greatly enhance emotional representation.

\begin{table}[t]
    \centering
    \caption{The quantitative comparison on the 3D-ETF. Best results in bold, second-best underlined.}
    \vspace{-5pt}
    \resizebox{\linewidth}{!}{
    \begin{tabular}{c|c|ccccc}
    \toprule
        \multirow{2}*{Dataset}  & \multirow{2}*{Method} & EVE $\downarrow$      & LVE $\downarrow$ & FDD $\downarrow$ & \multirow{2}*{FPS$\uparrow$} & \multirow{2}*{Size(MB)$\downarrow$}\\    
        \multirow{2}*{}         & \multirow{2}*{}       & ($\times 10^{-5}$)    & ($\times 10^{-5}$) & ($\times 10^{-7}$) \\
    \midrule
         \multirow{5}*{}  & FaceFormer & 2.487 & 8.656 & 11.909 & 426.01 & 349 \\
         \multirow{5}*{3D-ETF}  & EmoTalk & 1.331 &  2.347 & 2.196 & 1063.13 & 2411 \\
         \multirow{5}*{(Test)}  & FaceDiffuser & 2.537 & 6.137 & 8.172 & 10.33 & 860 \\
         \multirow{5}*{}  & SelfTalk & 2.206 & 6.796 & 11.622 & \underline{1428.57} & 1715 \\
         \multirow{5}*{}  & UniTalker-B & 1.439 & 1.727 & 3.004 & 68.75 & 348 \\
         \cmidrule{2-7}
         \multirow{5}*{}  & Ours-16 & \underline{1.275} & \textbf{1.021} & \textbf{2.023} & 440.01 & \underline{334} \\
         \multirow{5}*{}  & Ours-2 & \textbf{1.097} & \underline{1.375} & \underline{2.177} & \textbf{3632.15} & \textbf{45} \\
    \bottomrule
    \end{tabular}
    }
    \label{tab:eval_on_3detf}
   \vspace{-6mm}
\end{table}

\paragraph{Zero-shot evaluation.}
To test generalizability across all methods, we test models on VOCASET and calculate the LVE and FDD, as shown in Tab.~\ref{tab:eval_on_vocaset}.
Our approach demonstrates robust generalization capabilities in zero-shot scenarios and is highly competitive compared to state-of-the-art models.

\begin{table}[t]
    \centering
    \caption{The zero-shot test on the VOCASET.}
    \vspace{-5pt}
    \resizebox{0.8\linewidth}{!}{
    \begin{tabular}{c|c|ccc}
    \toprule
        \multirow{2}*{Dataset}  & \multirow{2}*{Method} & LVE $\downarrow$      & FDD $\downarrow$ & \multirow{2}*{Zero Shot}\\
        \multirow{2}*{}         & \multirow{2}*{}      & ($\times 10^{-5}$)    & ($\times 10^{-7}$) \\
    \midrule
         \multirow{5}*{VOCASET}  & FaceFormer &  1.170 & 2.493 & \ding{55}\\
         \multirow{5}*{(Test)}  & FaceDiffuser & 0.973 & 1.754 & \ding{55} \\
         \multirow{5}*{}  & SelfTalk & 0.967 & \textbf{1.049} & \ding{55} \\
         \multirow{5}*{}  & UniTalker-B & \textbf{0.814} & 1.396 & \ding{55} \\
         \multirow{5}*{}  & Ours-2 & \underline{0.857} & \underline{1.198} & \checkmark \\
    \bottomrule
    \end{tabular}
    } 
    \vspace{-4mm}
    \label{tab:eval_on_vocaset}
\end{table}

\paragraph{Ablation study.}
We conduct an ablation study to assess the impact of each component. In Tab.~\ref{tab:ablation_study}, the emotion embedding and enhancer have the greatest impact on EVE, while the enhancer's focus on the upper face leads to a slight degradation in lip region accuracy. In general, our distillation method enables the student model to learn effective knowledge from the teacher model, significantly improving model performance.

\begin{table}[t]
    \centering
    \caption{Ablation study for our 2-step model on 3D-ETF.}
    \resizebox{0.8\linewidth}{!}{
    \begin{tabular}{c|ccc}
    \toprule
        \multirow{2}*{Settings} & EVE$\downarrow$ & LVE $\downarrow$& FDD $\downarrow$ \\
        \multirow{2}*{}        & ($\times 10^{-5}$)    & ($\times 10^{-5}$) & ($\times 10^{-7}$) \\
    \midrule
          Ours-2 & \textbf{1.097} & 1.375 & \textbf{2.177}  \\
    \midrule
          w/o identity embedding & 1.223 & 1.412 & 2.301 \\
          w/o emotion embedding & 1.968 & 1.547 & 2.570 \\
          w/o enhancer & 1.712 & \textbf{1.116} & 2.512 \\
          w/o distillation & 1.305 & 1.647 &  2.618 \\
    \bottomrule
    \end{tabular}
    } 
    \vspace{-6mm}
    \label{tab:ablation_study}
\end{table}

\paragraph{Visualization comparison.}
Visualizations generated by the model are crucial for performance evaluation. We test all methods on BEAT-Test, 3D-ETF, and in-the-wild videos to compare facial animations. By feeding speech with diverse identities and emotions into the models, we show side-by-side comparisons with the ground truth (GT) in Fig.~\ref{fig:visualization}.
Our model demonstrates the highest sensitivity to varying identities. For example, in columns 1, 2, 5, and 6, when given audio from different speakers expressing the same emotion, it produces results closest to the GT, especially in the lip movements. When the same identity speaks with different emotions (columns 3, 4, 7, and 8), the model accurately reflects these emotional changes. Finally, when given speech with any identity and emotion, our model generates personalized facial animations, as seen in columns 9 and 10.

\section{Conclusion}

In this work, we introduced DiffusionTalker, a compact and efficient diffusion model to generate personalized 3D facial animations from speech. By combining identity and emotion embeddings and utilizing a personalizer-guided distillation approach, we significantly improve inference speed and reduce model size. The results show that DiffusionTalker delivers superior performance, making it suitable for real-time applications in virtual and augmented reality. This work provides a step forward in creating personalized and efficient models for interactive 3D facial animation.

\section{Acknowledgment}

This work was supported by the National Key R\&D Program of China (2024YFC3308500), the NSFC (62332015), the project of China Disabled Persons Federation (CDPF2023KF00002), and the Project of ISCAS (ISCAS-JCMS-202401 ).

\bibliographystyle{IEEEbib}
\bibliography{icme2025references}



\end{document}